\documentclass{article}
\usepackage[utf8]{inputenc}
\usepackage{natbib} 
\usepackage{graphicx}
\usepackage{todonotes}
\usepackage{lineno}
\usepackage{setspace}
\usepackage[affil-it]{authblk}
\usepackage{subcaption}
\usepackage{placeins}
\usepackage{amsmath}
\usepackage{hyperref}

\title{Identifying Wetland Areas in Historical Maps using Deep Convolutional Neural Networks}

\author{Niclas Ståhl \thanks{E-mail: \texttt{niclas.stahl@ju.se; Corresponding author}}}
\affil{Jönköping Artificial Intelligence Lab, \\ Jönköping University%, \\ Gjuterigatan 5, 553 18 Jönköping, Sweden
}
\affil{Skövde Artificial Intelligence Lab, \\ University of Skövde, \\ Kanikegränd 3, 541 34 Skövde,  Sweden}

\author{Lisa Weimann}
\affil{County Administrative Board of Jönköping%, \\ Länsstyrelsen i Jönköpings län, 551 86 Jönköping, Sweden
}

\date{\vspace{-5ex}}

\begin{document}

\maketitle

\clearpage

\begin{abstract}
    1) The local environment and land usages have changed a lot during the past one hundred years. Historical documents and materials are crucial in understanding and following these changes. Historical documents are, therefore, an important piece in the understanding of the impact and consequences of land usage change. This, in turn, is important in the search of restoration projects that can be conducted to turn and reduce harmful and unsustainable effects originating from changes in the land-usage.
    
    2) This work extracts information on the historical location and geographical distribution of wetlands, from hand-drawn maps.
    This is achieved by using \emph{deep learning (DL)}, and more specifically a \emph{convolutional neural network (CNN)}. The CNN model is trained on a manually pre-labelled dataset on historical wetlands in the area of Jönköping county in Sweden.
    These are all extracted from the historical map called ``Generalstabskartan''.
    
    3) The presented CNN performs well and achieves a $F_1$-score of 0.886 when evaluated using a 10-fold cross validation over the data. The trained models are additionally used to generate a GIS layer of the presumable historical geographical distribution of wetlands for the area that is depicted in the southern collection in Generalstabskartan, which covers the southern half of Sweden.
    This GIS layer is released as an open resource and can be freely used.
    
    4) To summarise, the presented results show that CNNs can be a useful tool in the extraction and digitalisation of non-textual information in historical documents, such as historical maps. A modern GIS material that can be used to further understand the past land-usage change is produced within this research. Previously, no material of this detail and extent have been available, due to the large effort needed to manually create such. However, with the presented resource better quantifications and estimations of historical wetlands that have been lost can be made.

\end{abstract}

\section{Introduction}

Historical maps hold crucial information about the landscape of the past, which is an important part in understanding ecological changes over time \citep{saar2012plant}.
Older historical maps are drawn by hand without any modern systems to aid, making the layout not fully consistent.
It is therefore a time consuming challenge to extract desired information from them.
The conventional approach to extract such information is through manual annotation, with the help of various GIS-software.
This labour intensive approach causes most current studies of historical landscapes and ecologies to be limited in size and only focusing on smaller areas or regions of particular interest, such as the study by \citet{cousins2009landscape}.

In some cases, automatic extraction of certain certain land covers can be done based on the colouring \citep{herrault2013automatic}.
The tool HistMapR \citep{auffret2017histmapr} is an example of software that has been proven useful for such automatic extraction.
However, the fading of colour and the yellowing of old paper is a disadvantage when analysing historical documents.
There are also historical maps that are drawn, or digitalised, in black and white and hence do not carry any colouring information.
The different land cover, therefore, need to be extracted by methods that analyses the different textures in the map, or discover different land cover implicitly, by analysing the surrounding landscape.

In this paper, we show how artificial intelligence (AI) can take advantage of previous manual annotation efforts that have been conducted.
More specifically, it is shown how a \emph{convolutional neural network (CNN)} can be trained using annotated data and thereafter be used to automatically detect areas of specific land cover in an old hand-drawn map.
As a proof of concept we show how a CNN can be trained to detect wetlands in the around one hundred years old \emph{Generalstabskartan} \citeauthor{generalstab}, which depicts the terrain and land-usage in Sweden.
The presented method is trained and evaluated on data from one county in Sweden (Jönköping county).
Besides this experiment, the model is also applied to historical maps covering the whole southern part of Sweden, with the aim of generating an overview of historical wetland areas. This overview can be used to quantify the historical wetland coverage. Such quantification enables analysis of the substantial loss of wetland area on a large scale. Hence such material is valuable when deciding where and how wetland restoration projects can be conducted.  
The result of the presented analysis, together with the source code for the method, is therefor released for public use.

CNNs are known to perform well at partitioning images into different segments, based on their content \citep{minaee2021image}.
Approaches similar to the one in the presented work have been used to study historical maps. For example,  \citet{saeedimoghaddam2020automatic} use a CNN to detect road intersections in historical maps provided by the United States Geological Survey (USGS).
Another study that uses DL to extract information from historical maps is \citet{weinman2019deep} who find and transcribe text from historical maps.
There are also some studies on more recent historical maps, such as \citet{le2020cnn} who segments historical orthoimages coupled to digital surface models (DSM) from the 1980s into different land cover.

Besides the analysis of historical maps, there are several applications utilising CNNs to more recent cartographic material.
One such application is the detection and segmentation of different land cover from multi-spectral remote sensing images \citep{huang2018urban}.
Beside detecting and classifying larger areas, CNNs have been used to detect very specific objects in remote sensing images and they can be applied with enough granularity to provide information about different species of trees \citep{branson2018google}.
If more granularity than satellite images can provide is needed, drones can be used to photograph an area.
The material, which is collected by the drone, can be analysed with the help of CNNs, and important objects can be detected and extracted.
Such approaches have been used to analyse and assess the ecological status of areas and also to detect and track different species \citet{gray2019drones,gray2019convolutional}.

All these cases highlight that CNNs can be useful in ecological applications that require visual analysis. 
In this work we show that a CNN based method can be applied to historical maps in order to extract information concerning the occurrence of wetlands.
Hence, the presented method can be used to minimise the manual labour that is required for analysing such data.
Using data from one region in southern Sweden, the County of Jönköping, we show that the method achieves both an high average precision, recall and $F_1$-score.

\section{Materials and Methods}

\subsection{Geographical area of analysis}
The presented analysis is conducted on the southern part of Sweden.
The limitation to just analyse the southern part arises from the earlier choice of drawing the southern part in a different scale than the northern part.
Furthermore, the model is trained on only one of the regions that the material covered.
The region that is selected is the Jönköping region, where pre-labeled data already existed.
In addition, this region covers several different nature types and a significant part of the region has historically been covered by wetlands.

\subsection{Convolutional neural networks}

The idea behind \emph{convolutional neural networks (CNNs)} was first presented by \citet{fukushima1982neocognitron} but the big breakthrough came some years later when \citet{krizhevsky2012imagenet} won the Imagenet competition, which is a competition focusing on object recognition in images.
CNNs are a special type of artificial neural networks and are inspired by the receptive fields of the human visual system.
The core strength of CNNs are their shift invariant property that allow them to detect a specific pattern in grid-like topologies, for example in an image, independently of the position of the said pattern.

\begin{figure}
    \centering
    \includegraphics[width=0.99\textwidth,height=0.99\textheight,keepaspectratio]{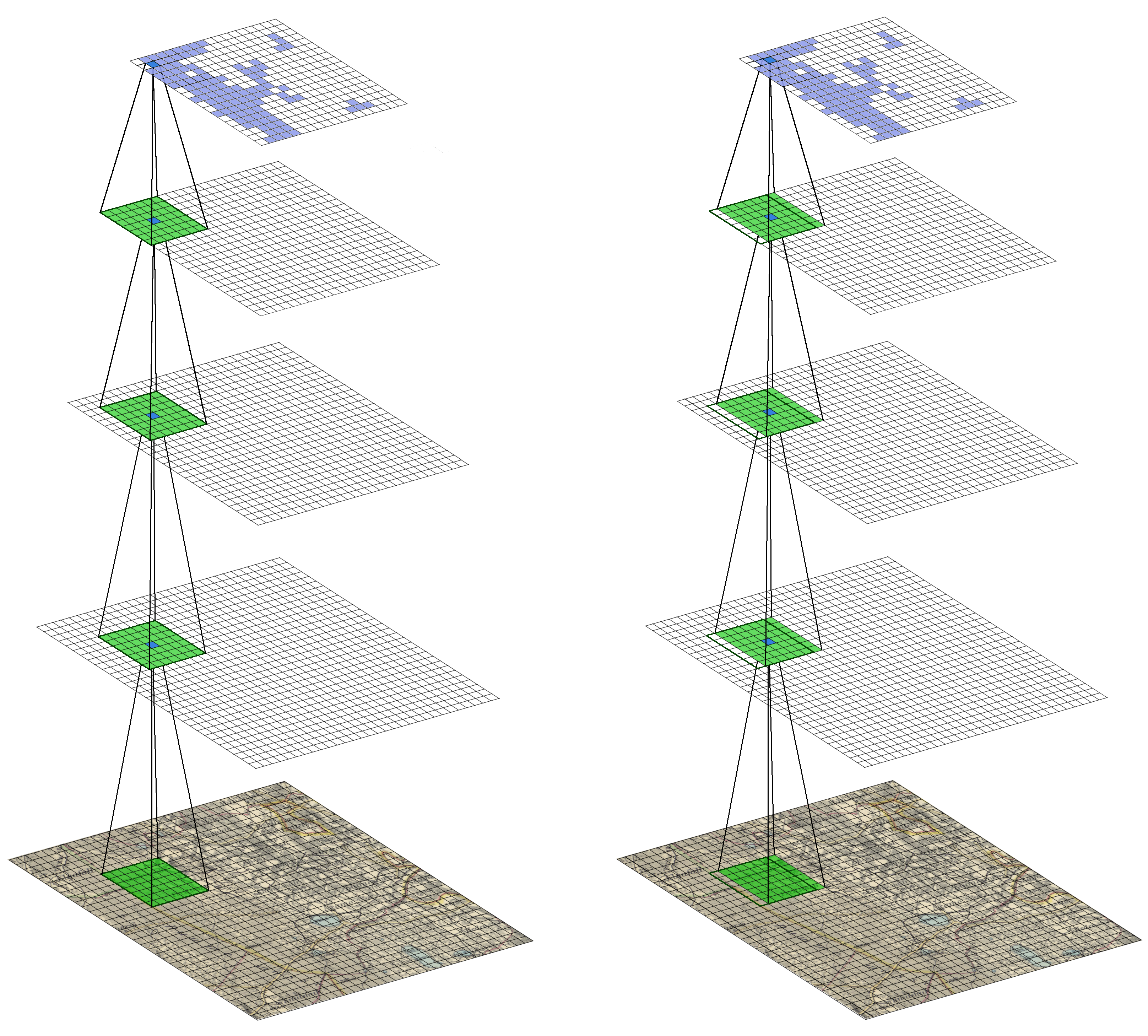}
    \caption{A CNN with four layers. A single pixel is highlighted with a dark blue colour in each of the layers. The values of all feature channels in this pixel are derived from the areas marked in green in the previous layers. In the very first layer the feature channels correspond to the RGB channels of the image, depicting the historical map.
    The right side of the figure shows the kernel being shifted one step to the right, compared to the left side. This shift results in the prediction of one pixel to the left in the final layer.
    }
    \label{fig:CNN}
\end{figure}

A CNN consists of several layers where, in each layer, a linear kernel is applied to a local area of the previous layer and by this produces a set of new latent features, as depicted in Figure \ref{fig:CNN}. Directly after the kernel a non-linear function is applied, in the presented research this is a leaky Rectified Linear Unit (ReLU) \citep{xu2015empirical}.
The kernel is repeatedly shifted over, and applied to, the previous representation creating a new grid-structure. An example of the whole process, when a CNN model is used to estimate the probability of a pixel in a historical map to be part of a wetland, is shown in Figure \ref{fig:CNN}. 

In the research literature it is common to include pooling layers after some of the activation functions. Such layers aggregate several neighbouring features into a single representation.
Hence, condensing the information of a larger area into a single representation.
A typical choice of pooling function is the \emph{max} or the \emph{sum} function.

\subsection{Approach}

As a proof of concept for the usefulness of deep learning, and more specifically CNNs, in the analysis of historical maps, this paper presents a case study where a historical map of one county in Sweden is analysed. A description of the map is presented in section \ref{sec:data}.
The CNN method presented in this paper is a full convolutional network, having 7 convolutional layers but no pooling layers. The full configuration of the network is presented in appendix \ref{app:architecture}.
The lack of pooling layers let the input signal flow directly from the input to the output, where the value of the pixel is determined.
Furthermore, no padding is added to the image, resulting in the loss of pixels close to the border of the image.
However, this is handled when the image segments are extracted from the full map, so that the overlap between the extracted images is large enough to make sure that no area of the map is missed out.

\begin{figure}
    \centering
    \begin{subfigure}[t]{0.3\textwidth}
		\centering
		\includegraphics[width=\textwidth]{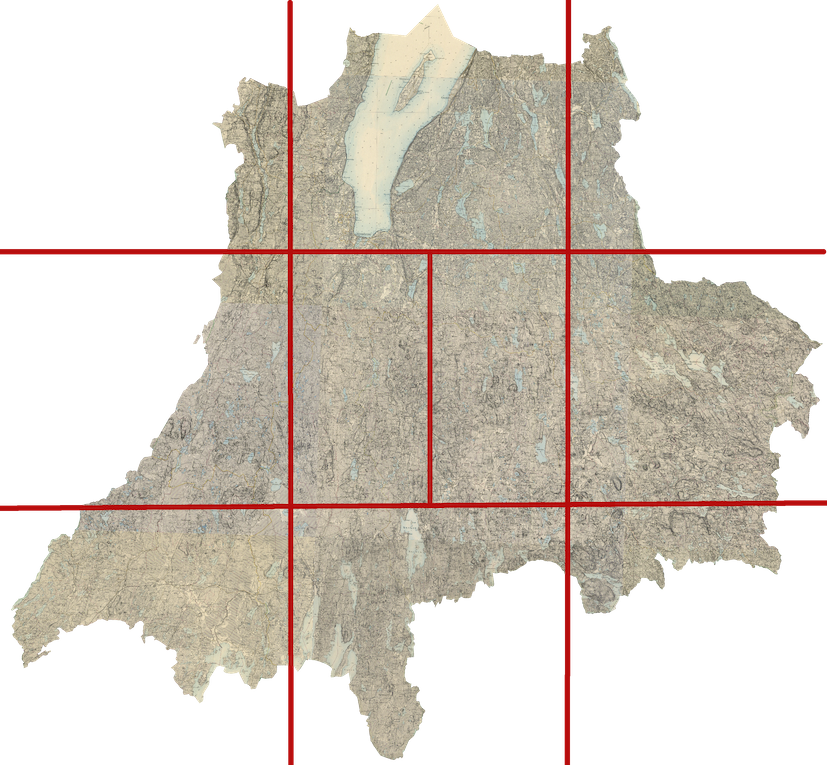}
		\caption{}
	\end{subfigure}
	~
	\begin{subfigure}[t]{0.3\textwidth}
		\centering
		\includegraphics[width=\textwidth]{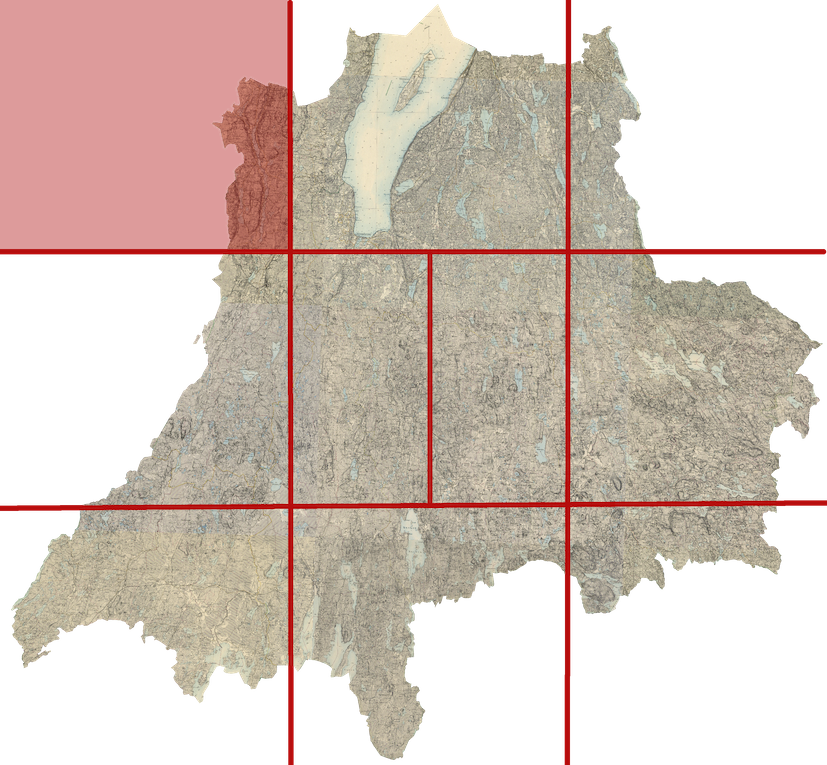}
		\caption{}
	\end{subfigure}
	~
	\begin{subfigure}[t]{0.3\textwidth}
		\centering
		\includegraphics[width=\textwidth]{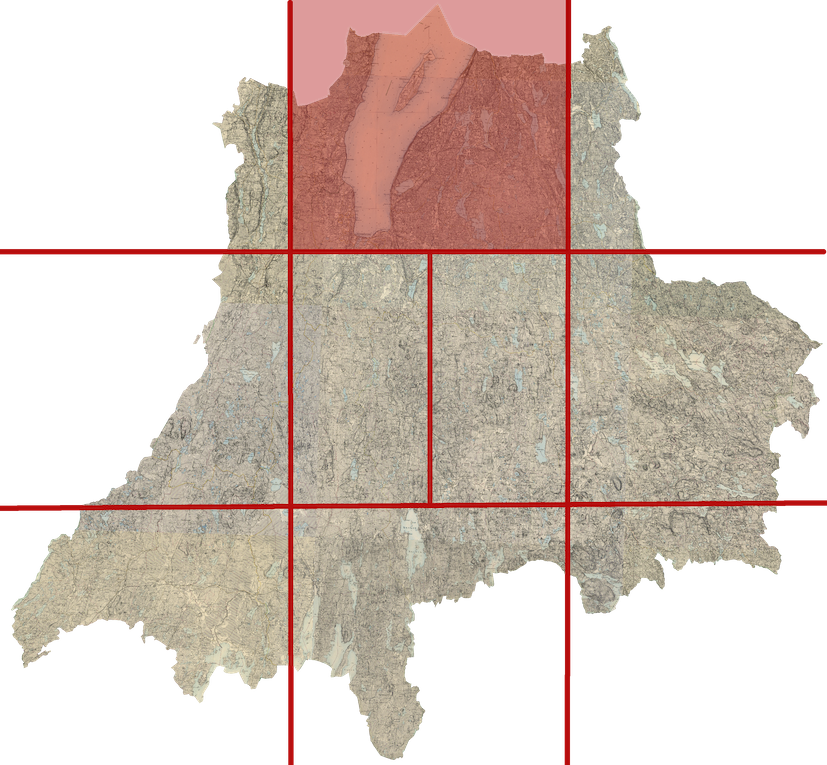}
		\caption{}
	\end{subfigure}
    \caption{Cross validation of the model. The model is cross validated in such a way that the analysed area is split into 10 sub-areas. These sub-areas varies in terrain type and hence, let us validate the generalisation behaviour of the model to areas with slightly different }
    \label{fig:CV}
\end{figure}

A 10-fold cross validation is performed in order for the result to be generalizable for the remaining maps, to which the CNN is also applied.
To create the 10 different sets for the cross validation, we split the map by placing a 3x3 grid over the map.
The region that is studies is not shaped as a square, and the central cell contains more area than the other 8.
This cell is, therefore, split into two cells making it 10 sets in total.
The division of the different sets are shown in figure \ref{fig:CV}.
During the training of the CNN 9 of these sets are used for the training and the final one is used for evaluation.
A challenge to the CNN is that the terrain differs in the different areas, as well as the style of the maps, and thus splitting the dataset in this way would give a good hint on the capability of the CNN to generalise.
Among the samples that are used for the training 20\% is used as a validation set to prevent the method from overfitting.
These samples are selected randomly from all areas that are used in the training set, and are never used to fit the model.
The model is trained for 150 epochs and with a batch size of 128. Dropout \citep{JMLR:v15:srivastava14a} with a rate of 0.3 is used during the training to make it more stable.
Furthermore, ADAM optimisation \citep{kingma2014adam} with a learning rate of $0.0001$ is used to find optimal weights in the neural network in order to minimise the cross entropy loss between the network's predictions and the pre-labelled data.

\subsection{Data pre-processing}

\label{sec:data}
As mentioned in the previous sections, the data is first split up into several larger blocks, depending on coordinates, with the purpose to cross validate the model.
These blocks are then split into many smaller areas of $80\times 80$ pixels, due to limitations in the available amount of memory. These splits are conducted in an iterative manner so the smaller areas are side by side to each other.
In addition, a padding of 27 pixels is added as a frame around the area used as input in order to counter the size reduction that occurs within the CNN.
This process creates 41601 smaller segments, which can be viewed as small images, that are all handled independently by the model.

\subsection{Data post-processing}

Some post-processing is required to transform the result of the CNN into an easy accessible GIS-resource.
This is primarily done to produce and refine the material covering southern Sweden, as well as making it easily accessible for further analyses.
This process consists of several steps.
In the first step, the pixel predictions from the CNN are rounded, so all predictions with predicted value larger than 0.5 are considered as wetlands and all predictions below are non-wetlands.
This creates a raster over the whole map, where each pixel is either deemed to be part of a wetland or not.
The next step is to convert this raster representation into a vector representation, to enable further analyses.
This conversion is also conducted to minimise storage space and making it easier to distribute.

In the final step, smaller wetlands, which are likely to arise due to noise and oddities in the map, are removed.
To achieve this, all wetlands that are less than $1000\text{m}^2$ are removed.

\subsection{Software and hardware}
All code that are used to produce the results in this paper is released as open source and is available at Github\footnote{\url{https://github.com/stan-his/GSK-NET}}
The model and all supporting programs are written in Python 3.7 and the model is implemented using PyTorch \citep{NEURIPS2019_9015}.
Furthermore, Rasterio \citep{gillies_2019} is used for the alignment of the map and the areas that are annotated as wetlands as well as for the rasterisation of the input.
Finally, Q-GIS and OpenStreetMap \citep{OpenStreetMap} are used to visualise the output and generate the output shown in Figure \ref{fig:Sweden}.

\section{Results}

\begin{figure}
    \centering
    \includegraphics[width=0.7\textwidth]{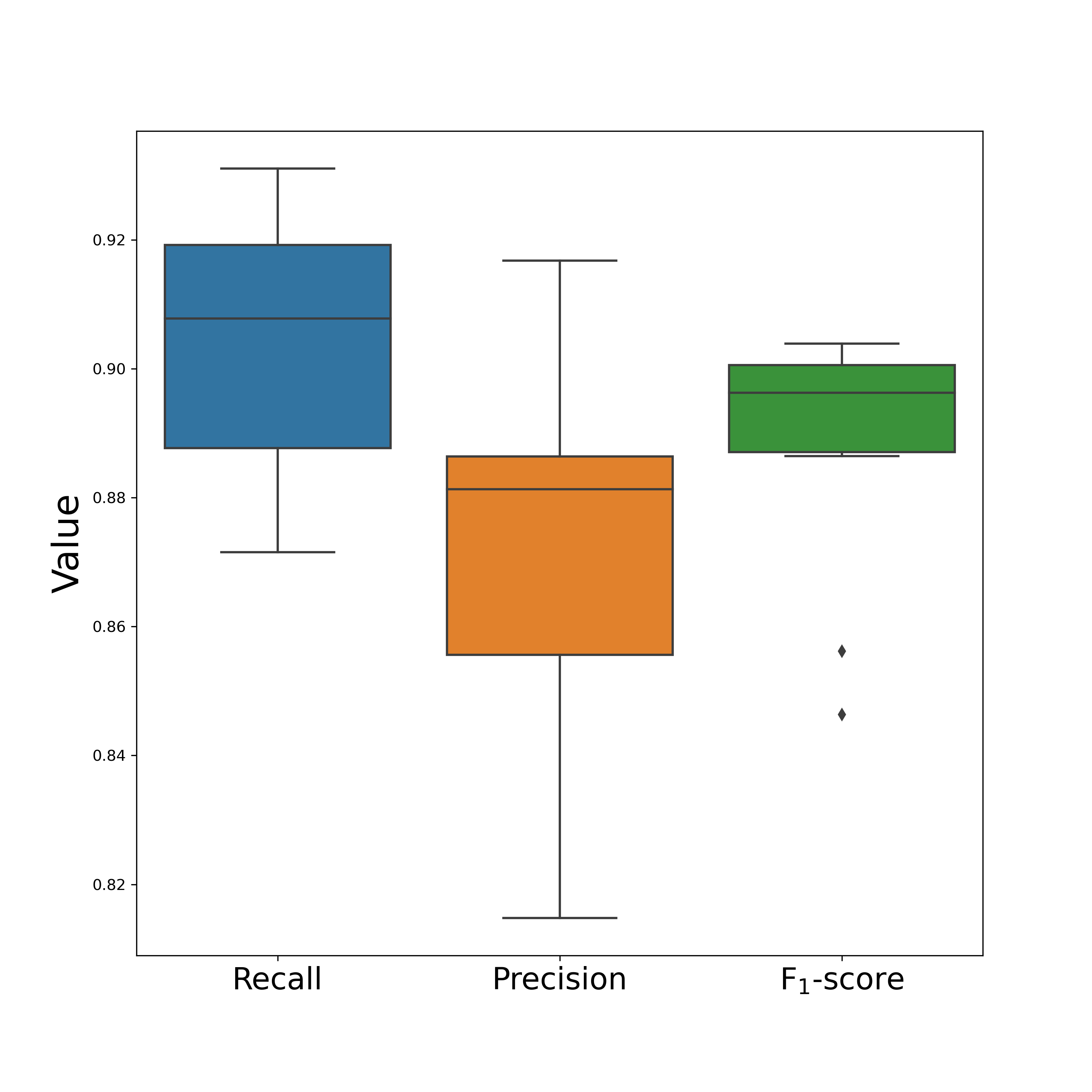}
    \caption{The distribution of precision, recall and $F_1$-score over the ten different folds.}
    \label{fig:resDist}
\end{figure}

\begin{figure}[h!]
	\centering
	\begin{subfigure}[t]{0.45\textwidth}
		\centering
		\includegraphics[width=\textwidth]{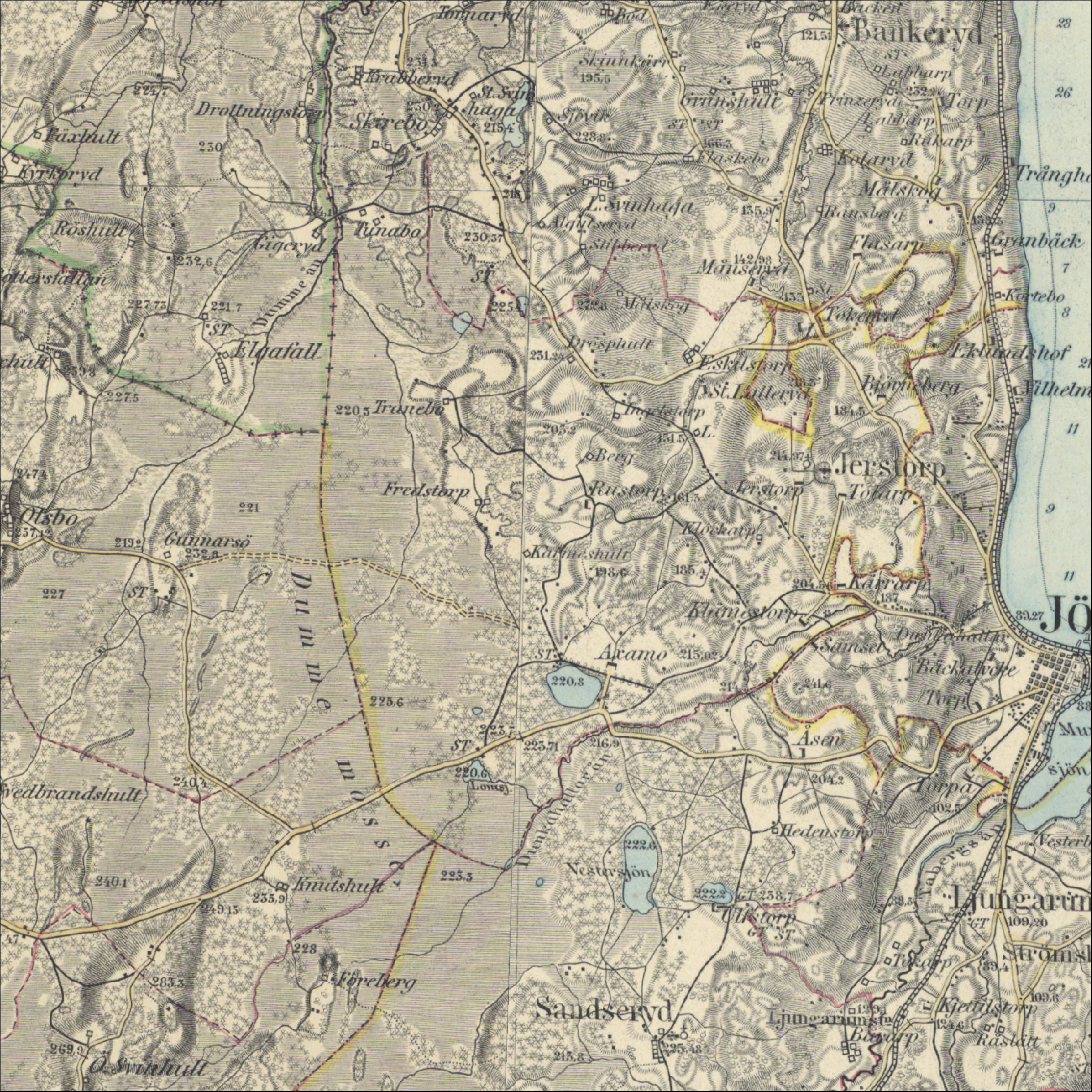}
		\caption{}
	\end{subfigure}
	~
    \begin{subfigure}[t]{0.45\textwidth}
		\centering
		\includegraphics[width=\textwidth]{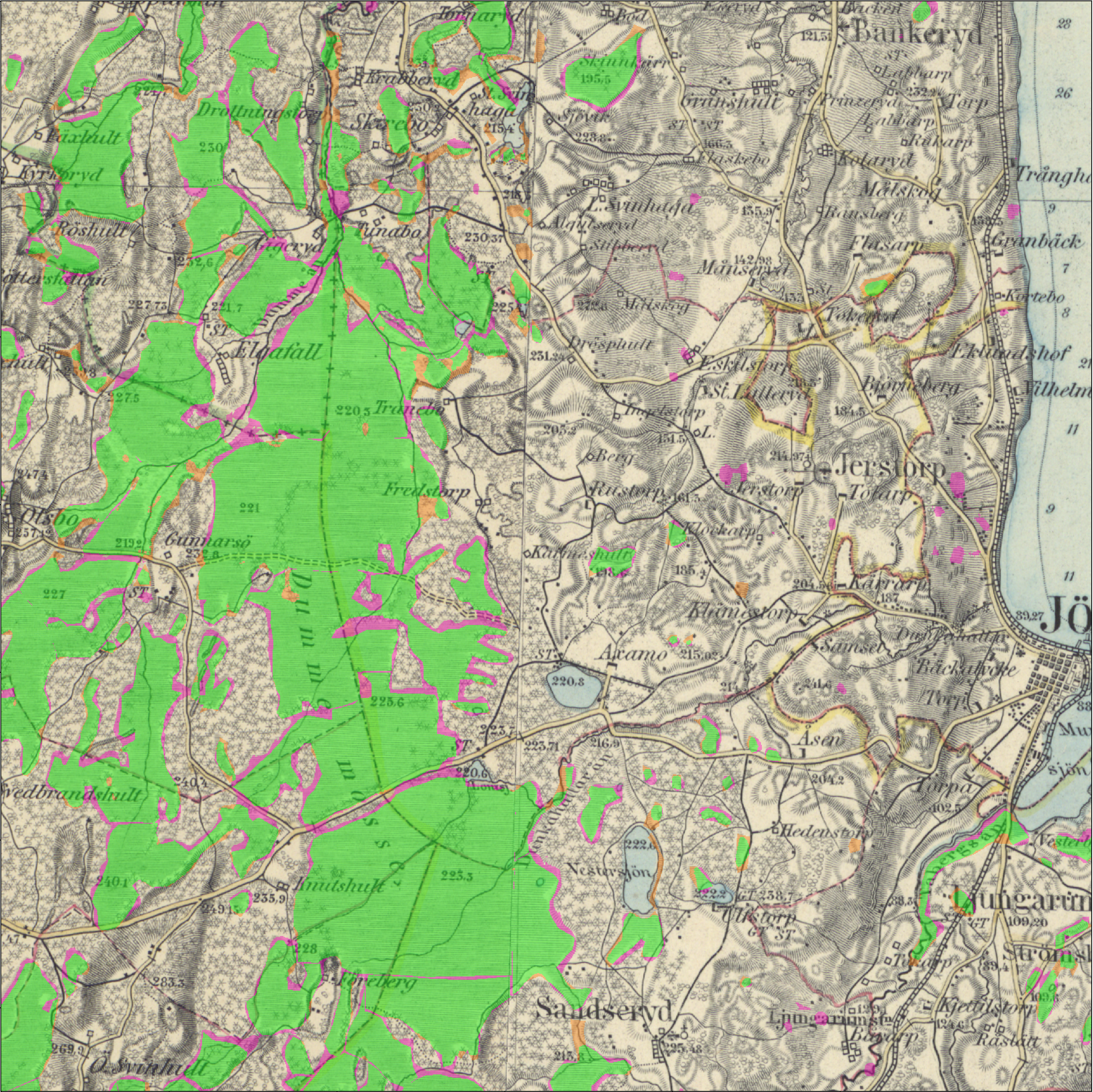}
		\caption{}
	\end{subfigure}
	
	\begin{subfigure}[t]{0.45\textwidth}
		\centering
		\includegraphics[width=\textwidth]{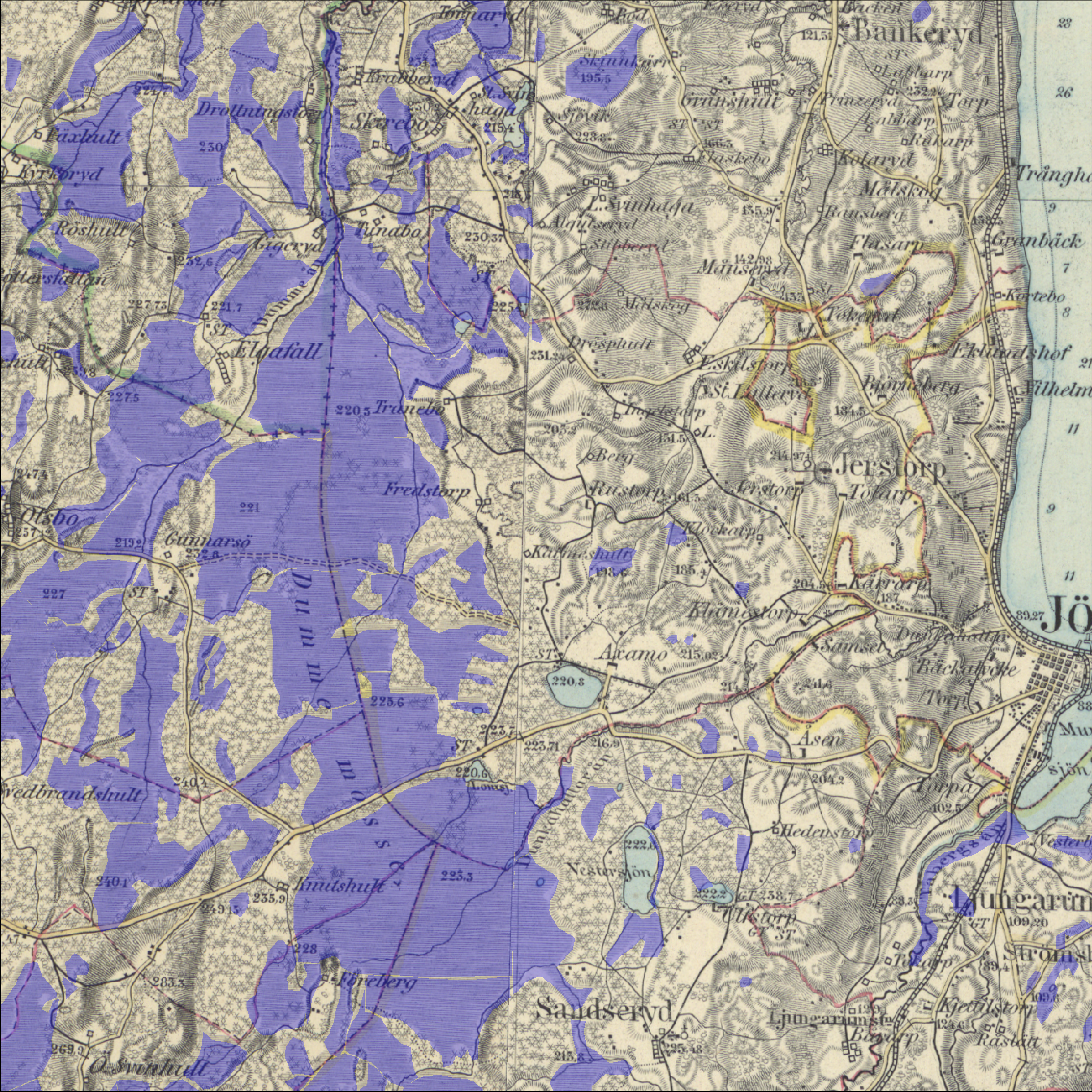}
		\caption{}
	\end{subfigure}
	~
    \begin{subfigure}[t]{0.45\textwidth}
		\centering
		\includegraphics[width=\textwidth]{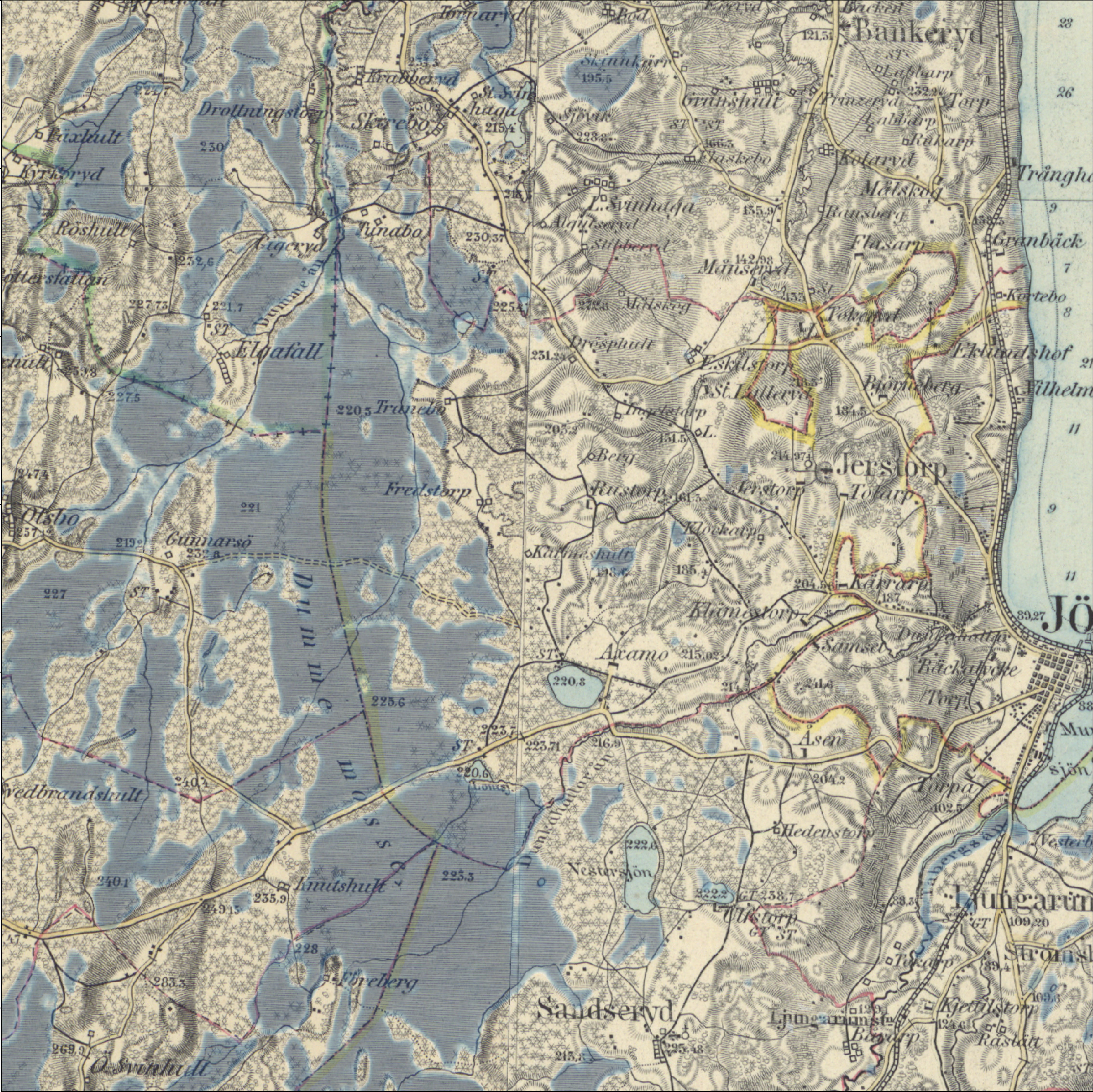}
		\caption{}
	\end{subfigure}
	\caption{An excerpt of the map is shown in (a). The areas that are annotated as wetlands, produced by a human, is coloured with blue and shown in (c).
	The corresponding annotation that is produced by a CNN, which has not seen this part of the map during training, is shown in (d).
	The similarities and differences between these two annotations are shown in (b). Here the areas for which the human and the CNN annotations are the same are displayed in green.
	The areas where the CNN annotate the land as wetland but the human did not (false positives) is displayed in pink.
	Finally, the areas that the humans annotated as wetland but, the CNN did not recognise as such (false negatives) are displayed in orange.
	}
	\label{fig:results}
\end{figure}

\begin{figure}
    \centering
    \includegraphics[width=0.75\textwidth]{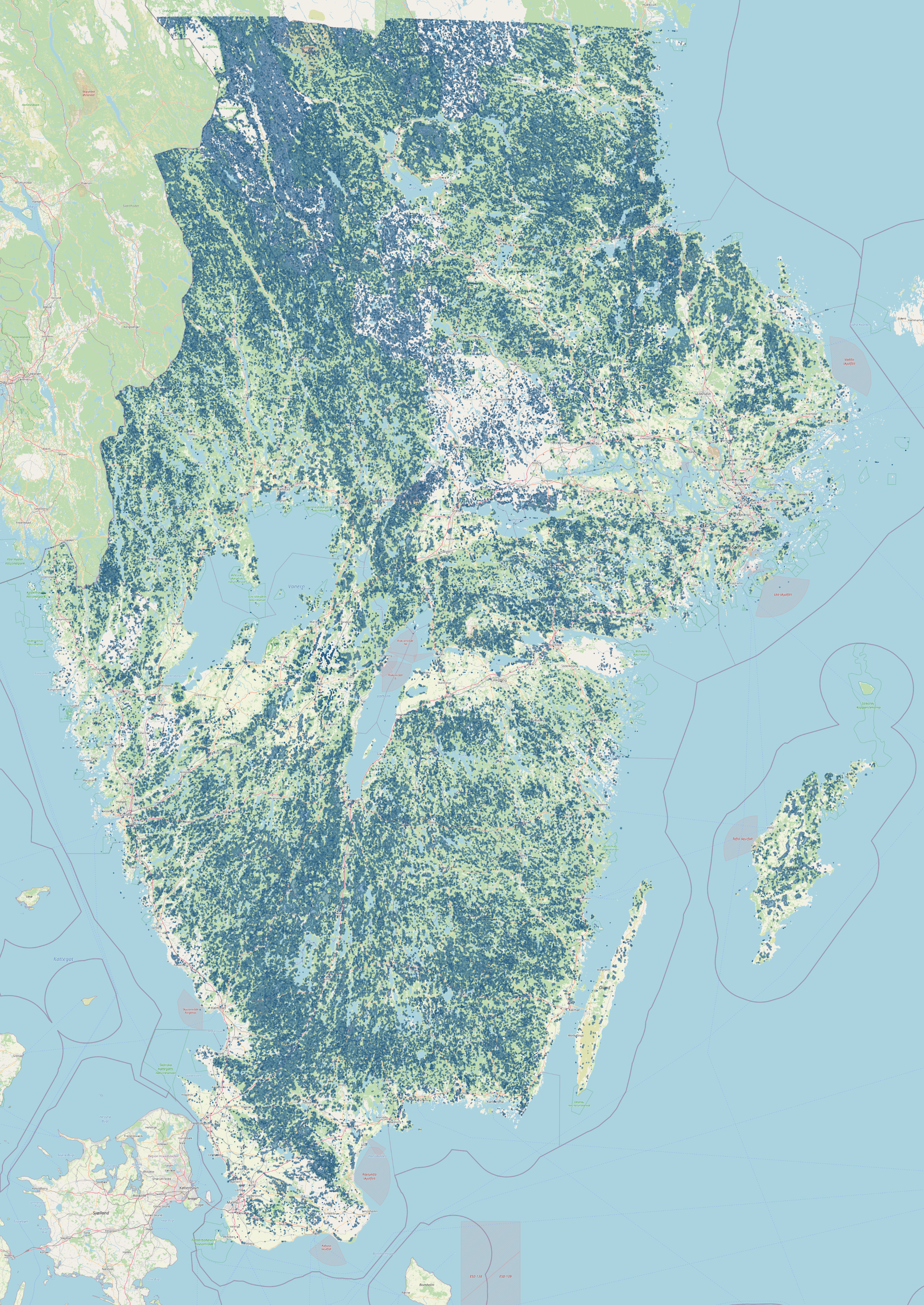}
    \caption{The CNN models estimation of wetlands in the southern part of Sweden, based on the map generalstabskartan. The background map is obtained from OpenStreetMap \citep{OpenStreetMap}.}
    \label{fig:Sweden}
\end{figure}

The presented method acquired a $F_1$-score, measured over all folds, of 0.886 where the precision of the model is 0.871 and the recall is 0.901.
The distribution over each of the different folds, of these three metrics, are shown in Figure \ref{fig:resDist}.
To further dissect how the model functions a smaller excerpt of the map and how the model classifies the different areas are shown in comparison with the annotated areas, in Figure \ref{fig:results}.
The total area that was covered by wetlands in the Jönköping region, when the studied maps were drawn, is estimated by the CNN to be $1.805*10^9\text{m}^2$.
This is an overestimation of the wetland area by 0.3\% compared to the area that is annotated by humans, which is $1.8*10^9\text{m}^2$.
When the same model is applied to the historical maps that cover the whole southern Sweden, the result of this is shown in Figure \ref{fig:Sweden}, the model estimates the total wetland area in the analysed area to have been $1.96*10^10\text{m}^2$ at the time that the map was drawn.
This can be compared to the modern-day wetland coverage in the analysed area of $7.82 * 10^9\text{m}^2$, which is provided by the Swedish Mapping, Cadastral and Land Registration Authority.
The results shown in Figure \ref{fig:Sweden} is published in Geodatakatalogen \footnote{\url{https://ext-geodatakatalog.lansstyrelsen.se/GeodataKatalogen/}}, a portal for sharing data that is hosted by the Swedish County Administrative Boards, and are available at no cost. 
Furthermore, it can be seen that areas that are known to be densely covered by wetlands today are also the areas which are most densely covered by wetlands in the models predictions, based on the historical map.
Another trend that can be spotted in the detected areas of wetland in the map is that areas that have historically been subject to agriculture contain few wetlands and have already been drained when the historical maps that were analysed were drawn.

\FloatBarrier

\section{Discussion}

Different AI methods are, more and more, used to reduce the workload that is currently needed to analyse different types material to form a basis for future decisions.
This paper explores how a CNNs can be used to detect and segment different land-usages from cartographic material. Several similar works has been been presented earlier with promising results, for example \citet{le2020cnn} and \citet{huang2018urban}.
However, most of these works are focused on modern multispectral images, most often extracted from remote sensing imagery, which hold less noise and have a coherent standard.
It is, however, shown in this paper that this type of approach that can be used, with promising performance, even for historical maps that are full of particularities and only follow a few standards.

The lack of pre-labelled material of high quality, which can be used in the training of supervised models, is a major bottleneck for full scale digitalisation of historical maps. 
In the presented case, it is shown that data from a single region, covering 173718 separate wetlands, where the smallest wetland is 3527$\text{m}^2$, is sufficient to get a model with acceptable performance.
However, no formal investigation about the amount of needed data is conducted and the amount of needed data is, of course, dependent on the studied nature type as well as the variability of the representations within the map.
Since the collection and annotation of data for the training of models, as the one in this paper, is a labour-intensive process, it would be valuable to perform such profound investigations and quantify the amount of data that may be needed.
Another way forward, which avoids the labour intense labelling, would be to look for already annotated data that have been used for other purposes and then use that data to build AI models. 
One additional solution, which needs to be further investigated, in order to avoid the collection of data is the generation of synthetic data from a smaller annotated dataset, using generative adversarial models, such as in the presented work by \citet{fang2019category} and \citet{li2019generating}.
However, even generative methods requires some annotated data to get started and it is uncertain if artefacts from the generative process will be kept in the generated data and how these artefacts will be expressed.

The presented method only considers the historical maps and no connection to the modern landscape is present. 
This may cause the generated information concerning the prevalence of wetlands to suffer from retrification errors as well as the preservation of artefacts and other flaws from the original map.
A future avenue for this research would, therefore, be to couple the generated information together with modern information, of much higher quality, such as soil and elevation data, following a similar approach to \citet{le2020cnn}.

\section{Conclusions}

The presented research in this paper shows that it is viable to extract environmental information from historical maps with the help of convolutional neural networks.
Our results shows that the performance of the CNN is on par with the human annotations and could minimise the burden of the manual digitalisation of historical maps.
The presented model achieves a $F_1$ score of 0.87 when a 10 fold cross validation is performed on the data.
The disagreement between the CNN and the pre-defined annotation can, furthermore, be explained by small disagreements on how the outline borders of the wetlands should be drawn.
This is supported by the fact that the agreement on a macroscopic level, where the agreement between the human annotator and the CNN is almost in unison.
In this case, the total area estimated by the CNN differed less than $0.3\%$ compared to the area that was marked by human annotators.

\subsection*{Acknowledgements}

First of all, we would like to thank Matti Ermold at the Swedish Environment Protection Agency who proposed the problem and guided us to relevant material and data and supported us during the project.
Secondly, we would like to thank Anne-Catrin Almér and Henrik Lindblom at the County Administrativ Board of Jönköping for the manual annotation of wetlands which form the data on which the model is trained.
Thirdly, We would like to thank Martin Axelsson and Robin Hellgren who explored possible models and worked with the targeted problem and data in their bachelor thesis at the University of Skövde.

\bibliography{references.bib}

\begin{thebibliography}{}

\bibitem[Auffret et~al., 2017]{auffret2017histmapr}
Auffret, A.~G., Kimberley, A., Plue, J., Sk{\aa}nes, H., Jakobsson, S.,
  Wald{\'e}n, E., Wennbom, M., Wood, H., Bullock, J.~M., Cousins, S.~A., et~al.
  (2017).
\newblock Histmapr: Rapid digitization of historical land-use maps in r.
\newblock {\em Methods in Ecology and Evolution}, 8(11):1453--1457.

\bibitem[Branson et~al., 2018]{branson2018google}
Branson, S., Wegner, J.~D., Hall, D., Lang, N., Schindler, K., and Perona, P.
  (2018).
\newblock From google maps to a fine-grained catalog of street trees.
\newblock {\em ISPRS Journal of Photogrammetry and Remote Sensing}, 135:13--30.

\bibitem[Cousins, 2009]{cousins2009landscape}
Cousins, S.~A. (2009).
\newblock Landscape history and soil properties affect grassland decline and
  plant species richness in rural landscapes.
\newblock {\em Biological Conservation}, 142(11):2752--2758.

\bibitem[Fang et~al., 2019]{fang2019category}
Fang, B., Kou, R., Pan, L., and Chen, P. (2019).
\newblock Category-sensitive domain adaptation for land cover mapping in aerial
  scenes.
\newblock {\em Remote Sensing}, 11(22):2631.

\bibitem[Fukushima and Miyake, 1982]{fukushima1982neocognitron}
Fukushima, K. and Miyake, S. (1982).
\newblock Neocognitron: A self-organizing neural network model for a mechanism
  of visual pattern recognition.
\newblock In {\em Competition and cooperation in neural nets}, pages 267--285.
  Springer.

\bibitem[Gillies et~al., 13  ]{gillies_2019}
Gillies, S. et~al. (2013--).
\newblock Rasterio: geospatial raster i/o for {Python} programmers.

\bibitem[Gray et~al., 2019a]{gray2019drones}
Gray, P.~C., Bierlich, K.~C., Mantell, S.~A., Friedlaender, A.~S., Goldbogen,
  J.~A., and Johnston, D.~W. (2019a).
\newblock Drones and convolutional neural networks facilitate automated and
  accurate cetacean species identification and photogrammetry.
\newblock {\em Methods in Ecology and Evolution}, 10(9):1490--1500.

\bibitem[Gray et~al., 2019b]{gray2019convolutional}
Gray, P.~C., Fleishman, A.~B., Klein, D.~J., McKown, M.~W., Bezy, V.~S.,
  Lohmann, K.~J., and Johnston, D.~W. (2019b).
\newblock A convolutional neural network for detecting sea turtles in drone
  imagery.
\newblock {\em Methods in Ecology and Evolution}, 10(3):345--355.

\bibitem[Herrault et~al., 2013]{herrault2013automatic}
Herrault, P.-A., Sheeren, D., Fauvel, M., and Paegelow, M. (2013).
\newblock Automatic extraction of forests from historical maps based on
  unsupervised classification in the cielab color space.
\newblock In {\em Geographic information science at the heart of Europe}, pages
  95--112. Springer.

\bibitem[Huang et~al., 2018]{huang2018urban}
Huang, B., Zhao, B., and Song, Y. (2018).
\newblock Urban land-use mapping using a deep convolutional neural network with
  high spatial resolution multispectral remote sensing imagery.
\newblock {\em Remote Sensing of Environment}, 214:73--86.

\bibitem[Kingma and Ba, 2014]{kingma2014adam}
Kingma, D.~P. and Ba, J. (2014).
\newblock Adam: A method for stochastic optimization.
\newblock {\em arXiv preprint arXiv:1412.6980}.

\bibitem[Krizhevsky et~al., 2012]{krizhevsky2012imagenet}
Krizhevsky, A., Sutskever, I., and Hinton, G.~E. (2012).
\newblock Imagenet classification with deep convolutional neural networks.
\newblock {\em Advances in neural information processing systems},
  25:1097--1105.

\bibitem[Le~Bris et~al., 2020]{le2020cnn}
Le~Bris, A., Giordano, S., and Mallet, C. (2020).
\newblock Cnn semantic segmentation to retrieve past land cover out of
  historical orthoimages and dsm: first experiments.
\newblock {\em ISPRS Annals of Photogrammetry, Remote Sensing and Spatial
  Information Sciences}, 2:pp--1013.

\bibitem[Li, 2019]{li2019generating}
Li, Z. (2019).
\newblock Generating historical maps from online maps.
\newblock In {\em Proceedings of the 27th ACM SIGSPATIAL International
  Conference on Advances in Geographic Information Systems}, pages 610--611.

\bibitem[Minaee et~al., 2021]{minaee2021image}
Minaee, S., Boykov, Y.~Y., Porikli, F., Plaza, A.~J., Kehtarnavaz, N., and
  Terzopoulos, D. (2021).
\newblock Image segmentation using deep learning: A survey.
\newblock {\em IEEE Transactions on Pattern Analysis and Machine Intelligence}.

\bibitem[{OpenStreetMap contributors}, 2017]{OpenStreetMap}
{OpenStreetMap contributors} (2017).
\newblock {Planet dump retrieved from https://planet.osm.org }.
\newblock \url{ https://www.openstreetmap.org }.

\bibitem[Paszke et~al., 2019]{NEURIPS2019_9015}
Paszke, A., Gross, S., Massa, F., Lerer, A., Bradbury, J., Chanan, G., Killeen,
  T., Lin, Z., Gimelshein, N., Antiga, L., Desmaison, A., Kopf, A., Yang, E.,
  DeVito, Z., Raison, M., Tejani, A., Chilamkurthy, S., Steiner, B., Fang, L.,
  Bai, J., and Chintala, S. (2019).
\newblock Pytorch: An imperative style, high-performance deep learning library.
\newblock In Wallach, H., Larochelle, H., Beygelzimer, A., d\textquotesingle
  Alch\'{e}-Buc, F., Fox, E., and Garnett, R., editors, {\em Advances in Neural
  Information Processing Systems 32}, pages 8024--8035. Curran Associates, Inc.

\bibitem[Saar et~al., 2012]{saar2012plant}
Saar, L., Takkis, K., P{\"a}rtel, M., and Helm, A. (2012).
\newblock Which plant traits predict species loss in calcareous grasslands with
  extinction debt?
\newblock {\em Diversity and Distributions}, 18(8):808--817.

\bibitem[Saeedimoghaddam and Stepinski, 2020]{saeedimoghaddam2020automatic}
Saeedimoghaddam, M. and Stepinski, T.~F. (2020).
\newblock Automatic extraction of road intersection points from usgs historical
  map series using deep convolutional neural networks.
\newblock {\em International Journal of Geographical Information Science},
  34(5):947--968.

\bibitem[Srivastava et~al., 2014]{JMLR:v15:srivastava14a}
Srivastava, N., Hinton, G., Krizhevsky, A., Sutskever, I., and Salakhutdinov,
  R. (2014).
\newblock Dropout: A simple way to prevent neural networks from overfitting.
\newblock {\em Journal of Machine Learning Research}, 15(56):1929--1958.

\bibitem[{Swedish Mapping, Cadastral and Land Registration Authority},
  2021]{generalstab}
{Swedish Mapping, Cadastral and Land Registration Authority} (2021).
\newblock Geographical survey office archive.

\bibitem[Weinman et~al., 2019]{weinman2019deep}
Weinman, J., Chen, Z., Gafford, B., Gifford, N., Lamsal, A., and Niehus-Staab,
  L. (2019).
\newblock Deep neural networks for text detection and recognition in historical
  maps.
\newblock In {\em 2019 International Conference on Document Analysis and
  Recognition (ICDAR)}, pages 902--909. IEEE.

\bibitem[Xu et~al., 2015]{xu2015empirical}
Xu, B., Wang, N., Chen, T., and Li, M. (2015).
\newblock Empirical evaluation of rectified activations in convolutional
  network.
\newblock {\em arXiv preprint arXiv:1505.00853}.

\end{thebibliography}

\appendix
\section{Network architecture}
\label{app:architecture}

\FloatBarrier

\begin{table}[h!]
    \centering
    \begin{tabular}{|r l|}
    \hline
    Hidden layers & = 6\\
    Neurons & = 128, 64, 64, 32, 32, 32\\
    Kernel sizes & = 9x9, 9x9, 7x7, 7x7, 7x7, 5x5, 5x5\\
    Dropout rate & = 0.3 \\
    Optimiser & = Adam \\
    Learning rate & = 0.0001 \\
    \hline
    \end{tabular}
    \caption{Parameters for the CNN model that is used.}
    \label{tab:appPara}
\end{table}

\end{document}